\documentclass[lettersize,journal]{IEEEtran}

\usepackage{amsmath,amsfonts,amssymb}
\usepackage{algorithmic}
\usepackage{algorithm}
\usepackage{array}
\usepackage[caption=false,font=normalsize,labelfont=sf,textfont=sf]{subfig}
\usepackage{textcomp}
\usepackage{stfloats}
\usepackage{url}
\usepackage{verbatim}
\usepackage{graphicx}
\usepackage{cite}
\usepackage{booktabs}
\usepackage{multirow}
\usepackage[table]{xcolor}

\newcommand{\cmark}{\ensuremath{\checkmark}}

\graphicspath{{./}}
\def\BibTeX{{\rm B\kern-.05em{\sc i\kern-.025em b}\kern-.08em
    T\kern-.1667em\lower.7ex\hbox{E}\kern-.125emX}}

\title{Boosting Infrared Small Target Detection via Logit-Domain Contrast and Adaptive Shape Refinement}

\author{Handong Zeng,
    Zhengeng Yang*,
    Shuai Zhang,
    Shikai Chen,
    and~Hongshan Yu%
\thanks{Handong Zeng, Zhengeng Yang, Shuai Zhang, and Shikai Chen are with the College of Engineering and Design, Hunan Normal University, Changsha 410081, China (Corresponding author: Zhengeng Yang. email: \textcolor{magenta}{zhdydkdh@163.com, yzg050215@163.com, zshuai2004@163.com, 775862465@qq.com}).}
\thanks{Hongshan Yu is with the School of Artificial Intelligence and Robotics, Hunan University, Changsha 410082, China (email: \textcolor{magenta}{yuhongshancn@hotmail.com}).}}

\IEEEaftertitletext{%
\noindent
\begin{minipage}[t]{0.48\textwidth}
    \vspace{0pt}
    \small\bfseries\textit{Abstract}---Infrared small target detection (IRSTD) remains challenging due to tiny target size, low signal-to-noise ratio, severe foreground-background imbalance, and blurred boundaries in complex scenes. Existing methods usually rely on post-activation probability-domain supervision for discrimination, where weak targets and strong clutter may produce saturated and close probabilities, limiting weak-target discrimination. Meanwhile, blurred boundaries and halo-like predictions mainly stem from thermal diffusion, tiny target scale, boundary uncertainty, and insufficient explicit contour constraints. To address these issues, we propose Adaptive-Contrastive SLSIoU (AC-SLSIoU), a plug-and-play discriminative and shape-aware loss for IRSTD. Specifically, a Logit-Domain Margin Constraint (LDMC) is introduced to enlarge the response gap between targets and informative hard negatives in the logit space, thereby enhancing weak-target discrimination. Adaptive Boundary Suppression (ABS) applies scale-aware annular penalties to refine target contours and suppress halo-like overflow responses. In addition, False-Alarm Focal Loss assigns larger weights to high-probability negative samples, further penalizing persistent high-confidence false alarms. Without introducing extra inference overhead, the proposed method can be seamlessly integrated into existing detectors and consistently improves both detection accuracy and shape quality. Extensive experiments and cross-backbone evaluations demonstrate the effectiveness, robustness, and generalization ability of the proposed method for infrared small target detection.

    \vspace{1.34ex}
    \textit{Index Terms}---Infrared small target detection, logit-domain margin constraint, adaptive boundary suppression, false-alarm focal loss.
\end{minipage}\hfill
\begin{minipage}[t]{0.48\textwidth}
    \vspace{0pt}
    \centering
    \includegraphics[width=\linewidth]{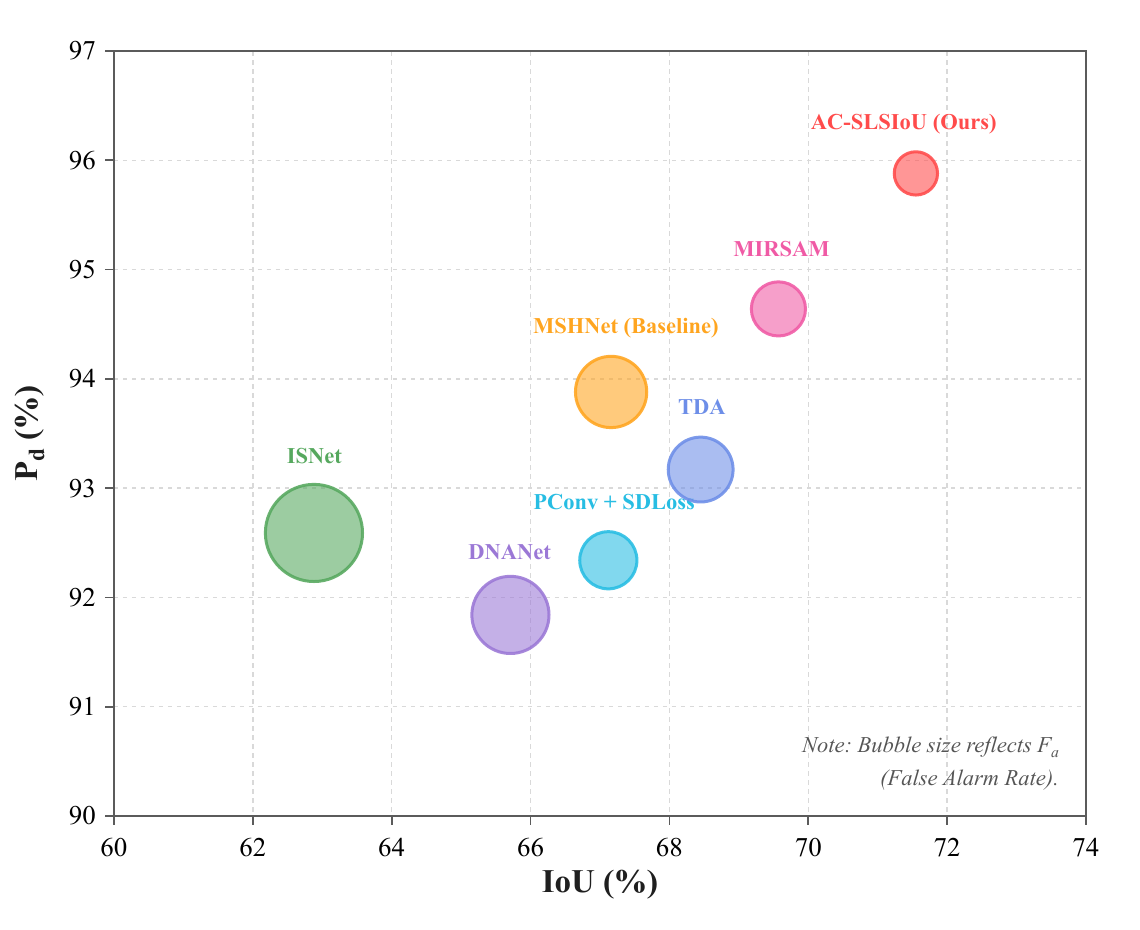}
    \refstepcounter{figure}\label{fig:performance_distribution}
    \vspace{0.35ex}

    {\footnotesize Fig.~\thefigure. Performance distribution of the compared methods.\par}
\end{minipage}
\vspace{1.0\baselineskip}
}

\begin{document}

\maketitle

\IEEEpeerreviewmaketitle

\section{Introduction}
\IEEEPARstart{I}{nfrared} Small Target Detection (IRSTD)~\cite{liu2024sls,zhang2025saist,mirsam2025,yang2025pconv,yang2025infraredsurvey} plays a vital role in remote surveillance and early warning systems. Unlike generic object detection in optical images, IRSTD faces a series of extreme challenges: targets typically occupy only a few pixels, lack texture and color information, and are often submerged in heavy background clutter and high-intensity noise. Consequently, the core objective of IRSTD is to achieve pixel-level accuracy---specifically, maximizing the detection rate of weak targets while ensuring high Intersection over Union (IoU) through precise shape estimation. 

In recent years, deep learning-based methods have made significant progress in this field by learning data-driven features. Although existing convolutional neural networks excel in feature extraction, most detectors rely on loss functions constructed in the post-activation probability domain during supervised training, such as SoftIoU~\cite{rahman2016iou} and Binary Cross-Entropy (BCE)~\cite{ronneberger2015unet}. While these loss functions are effective for general segmentation tasks, they often encounter bottlenecks when dealing with the extreme class imbalance and low signal-to-noise ratio characteristics inherent in IRSTD. Specifically,  when distinguishing extremely weak targets from bright background clutter (such as cloud edges), their probability values in the bounded [0,1] probability space may be very close to each other. Such tiny numerical differences generate only weak gradient signals, hindering the model from mining deep discriminative features and leading to missed detections of weak targets. 

Besides the gradient saturation problem, another issue ignored by existing methods is the shape blurring problem. Due to the thermal radiation effect, infrared detection tends to produce a halo phenomenon around the target. However, existing methods typically did not impose explicit constraints on this issue and fail to obtain accurate target boundaries, i.e., the prediction results retains the blurred ``halo effect'' introduced by the input, which severely restricts the segmentation performance.

To address these two issues mentioned above, we reformulate the training objective from two complementary directions. To tackle the gradient saturation problem, we add a supervision signal from the logit domain. For the ``halo effect'' issue, a scale-adaptive boundary penalty is introduced to explicitly constrain the target contours. Based on these two ideas, our contributions can be summarized as follows

\begin{itemize}
\item \textbf{Logit-Domain Margin Constraint (LDMC):} 
Inspired by the mechanism of contrastive learning that pushes positive and negative samples apart, we first presented a  Logit-Domain Margin Constraint loss, which introduces an explicit constraint on the response values of targets and backgrounds in the logit domain. Specifically, this loss is designed by constructing target–hard-negative sample pairs and adding a distance constraint on the logit-domain response between each pair, which prevents the issue that overly similar probability-domain outputs between the paired samples failing to provide meaningful supervisory signals.

%Motivated by logit adjustment strategies in long-tailed learning~\cite{chen2025robust} , we first impose a margin constraint for the logit values between the targets and backgrounds before Sigmoid activation. . constructs pixel-level target-clutter pairs in the logit domain, selects high-response background pixels as hard negatives, and enforces an explicit margin between target and clutter logits to strengthen weak-target discrimination under probability-domain saturation;
\item \textbf{Adaptive Boundary Suppression (ABS):} To address halo-like shape blurring caused by thermal diffusion, we then proposed Adaptive Boundary Suppression loss. The core idea of ABS is to generate an annular penalty region around the target whose size adapts to the target scale, and suppress false detections falling within this region, so as to eliminate the interference caused by halos around the input target. 

\item \textbf{False-Alarm Focal Loss (FAFL)} We observe that numerous stubborn false alarms remain even after employing the above two losses. Considering that these false alarms obtain exceedingly high prediction probabilities over the positive classes, inspired by Focal Loss which focuses on hard samples with low prediction probabilities, we introduce an False-Alarm Focal Loss. It assigns higher weights to negative samples with high predicted probabilities over the positive class.

%To complement the logit-domain margin and boundary constraints, a lightweight focal suppression term is activated for persistent high-confidence background responses. This term focuses supervision on hard clutter and improves false-alarm suppression without introducing inference overhead.
\end{itemize}

To verify the effectiveness of the proposed strategy, we selected the MSHNet as our baseline, which introduced only a Scale and Location Sensitive (SLS) loss to the commonly used UNet-based segmentation architecture.  We added our three losses to the basic SLS loss and named the total loss as AC-SLSIoU. As illustrated in Fig.~\ref{fig:performance_distribution}, our AC-SLSIoU is located in the favorable region with higher IoU and $P_d$ and a smaller bubble size corresponding to lower $F_a$, indicating a better balance among segmentation accuracy, target detection, and false-alarm suppression than the baseline and other compared methods. Experiments on the IRSTD-1k and NUDT-SIRST datasets further demonstrate that this method improves detection accuracy and shape quality without increasing inference costs.

\section{Related Work}

\subsection{Infrared Small Target Detection}

\textbf{Traditional approaches.} Early IRSTD methods mainly relied on manually designed priors, and their development generally follows the increasing complexity of background modeling. The first line uses local filtering to remove slowly varying backgrounds, where Top-Hat~\cite{deng2022entropy} and Max-Median~\cite{deshpande1999maxmedian} enhance small bright responses by estimating surrounding background statistics. These methods are simple and efficient, but they are sensitive to high-intensity clutter and noise. To better exploit the saliency difference between targets and their neighborhoods, local contrast methods such as LCM~\cite{chen2014lcm}, WSLCM~\cite{han2021wslcm}, and TLLCM~\cite{han2020tllcm} model target-background contrast within hand-crafted windows. They improve target enhancement in relatively clean scenes, yet their fixed local assumptions often fail when cloud edges, sea clutter, or man-made heat sources show similar contrast patterns. A more global line formulates the infrared image as a combination of low-rank background and sparse targets. IPI~\cite{gao2013ipi}, RIPT~\cite{dai2017ript}, PSTNN~\cite{zhang2019pstnn}, and MSLSTIPT~\cite{sun2021mslstipt} improve robustness by exploiting nonlocal correlations or tensor structures. However, these optimization-based methods usually involve iterative solvers, making real-time deployment difficult.

\textbf{Deep learning architectures.} Deep learning methods first moved IRSTD from hand-crafted priors to data-driven feature learning. Context- and attention-based detectors such as ACM~\cite{dai2021acm} and ALCNet~\cite{dai2021alcnet} introduce asymmetric context modeling and local contrast attention to strengthen target cues. Dense aggregation methods such as DNANet~\cite{li2022dnanet} further improve multi-level feature reuse through nested interactions. Since infrared targets are not only small but also shape-sensitive, ISNet~\cite{zhang2022isnet} introduces shape-aware constraints, and MSHNet~\cite{liu2024sls} incorporates scale- and location-sensitive supervision to improve mask quality.

Following these early CNN-based designs, recent methods can be grouped according to the specific limitations they address. To improve feature fusion under complex backgrounds, RRCANet~\cite{liu2025rrcanet} refines target semantics with recurrent reusable convolutions and attention aggregation, NFPN\&IS~\cite{lu2025nfpnis} combines nested feature pyramids with interference suppression, MLP-Net~\cite{wang2025mlpnet} explores MLP-based global interaction, MTMLNet~\cite{yang2025mtmlnet} introduces multi-task mutual learning, Diverse Feature Harmonization~\cite{guo2025dfh} aligns heterogeneous feature responses, Multibranch Mutual-Guiding Learning~\cite{li2025mml} promotes complementary branch interaction, Local-Global Feature Fusion~\cite{wu2025localglobal} combines local details with global context, and Triple-Directional Fusion Attention~\cite{wang2025tdfa} strengthens directional feature aggregation. Another group focuses on frequency-domain, physical, or dynamic modeling. PConv + SDLoss~\cite{yang2025pconv} enhances small-target feature extraction with pinwheel-shaped convolution and scale-based dynamic loss, Computational Fluid Dynamic Network~\cite{zhang2025cfdnet} introduces fluid-inspired dynamic modeling, Structure-Guided Diffusion Filter~\cite{zhang2025sgdf} uses structural diffusion to suppress background interference, FADet~\cite{feng2025fadet} models frequency-aware detection cues, Dynamic High-Frequency Convolution~\cite{li2026dhc} emphasizes high-frequency target structures, and Spatial--Frequency Feature Learning~\cite{li2026sffl} jointly exploits spatial and spectral representations. Beyond CNN-centric designs, SAMamba~\cite{xu2025samamba} and Spectrum-Assisted Mamba~\cite{li2025spectrum} introduce state-space modeling for long-range dependency capture, while MIRSAM~\cite{mirsam2025} adapts the Segment Anything framework with multimodal language guidance. Specialized and efficient variants have also emerged: DISTA-Net~\cite{han2025dista} addresses closely spaced targets through dynamic unmixing, DSAD~\cite{li2025dsad} performs feature distillation with multi-directional contrast spatial attention, and Frequency-Aware Lightweight Distillation~\cite{cai2026fald} explores lightweight teacher-student transfer.

Overall, existing deep IRSTD methods have progressively improved feature representation from local context modeling to global interaction, frequency modeling, state-space modeling, and foundation-model adaptation. Nevertheless, most of them still optimize predictions mainly in the post-activation probability domain. In contrast, this paper focuses on reformulating the supervision objective itself by introducing logit-domain margin constraints and adaptive shape refinement, thereby enhancing discrimination and contour quality without changing the inference architecture.

\subsection{Logit-Domain Margin Constraint}

The motivation of logit-domain supervision is related to two research directions: logit adjustment and contrastive discrimination. In long-tailed recognition, Robust Logit Adjustment~\cite{chen2025robust} shows that reshaping pre-activation decision scores can alleviate class imbalance more directly than only reweighting probabilities. This observation is relevant to IRSTD, where the foreground occupies only a tiny fraction of pixels and the target response is easily compressed after sigmoid activation.

Contrastive learning provides another perspective for enhancing separability. MoCo~\cite{he2020moco} learns discriminative visual representations by pulling positive samples together and pushing negative samples apart in a normalized feature space. For IRSTD, directly applying instance-level contrast is difficult because weak targets and hard clutter often differ only in subtle pixel-level responses. Recent infrared-oriented methods begin to exploit target-background separation more explicitly. SAIST~\cite{zhang2025saist} uses contrastive language-image guidance to improve target discrimination, and WeCoL~\cite{duan2026wecol} introduces weakly supervised contrastive learning with quantity prompts to distinguish moving infrared small targets. Loss-level methods such as TDA~\cite{tda2025} and PConv + SDLoss~\cite{yang2025pconv} also adjust supervision according to target difficulty or scale, but they still mainly operate through probability-domain weighting.

Different from the above methods, the proposed LDMC transfers the contrastive idea from feature or prompt spaces to the raw logit response space. Instead of comparing global representations, it constructs pixel-level target-clutter pairs and enforces an explicit margin before sigmoid activation. This design provides continuous gradients when target and background probabilities are already saturated but their logits remain insufficiently separated, making it suitable for weak-target discrimination under severe foreground-background imbalance.

\subsection{Boundary Refinement}

Accurate boundary estimation is another key issue in IRSTD because thermal diffusion often produces halo-like predictions around small targets. Existing methods address this problem from different perspectives. ISNet~\cite{zhang2022isnet} emphasizes that shape information is essential for infrared small target segmentation, MSHNet~\cite{liu2024sls} improves localization through scale- and location-sensitive supervision, TDA~\cite{tda2025} enhances low-contrast targets by target-driven adaptive weighting, and PConv + SDLoss~\cite{yang2025pconv} uses scale-based dynamic loss to stabilize small-target learning. Structure-Guided Diffusion Filter~\cite{zhang2025sgdf} further introduces structural guidance to reduce background interference. These methods improve contour quality or localization indirectly through shape priors, scale weighting, or feature filtering.

However, generic boundary-aware losses and fixed morphological penalties may be mismatched with the extreme scale variation of infrared targets. A fixed boundary band can over-suppress tiny targets while under-constraining larger targets. Therefore, our Adaptive Boundary Suppression mechanism explicitly constructs a scale-aware penalty ring outside each target. By suppressing responses in this adaptive surrounding region, it directly reduces halo-like over-segmentation while preserving target pixels, leading to more compact contours and improved pixel-level IoU.

\section{Methodology}
\begin{figure*}[!t]
  \centering
    \includegraphics[width=\textwidth]{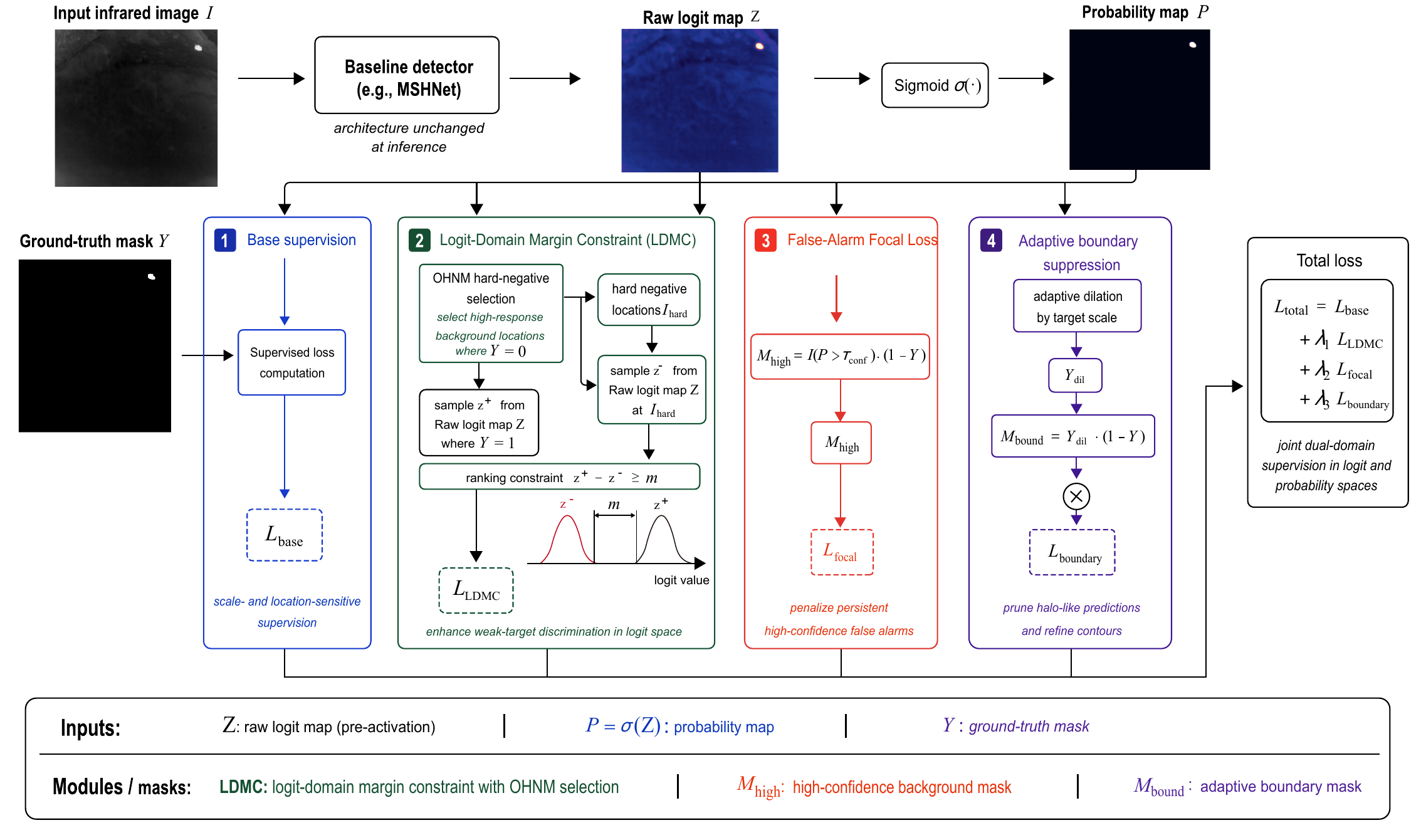}
  \caption{Given an infrared image, the backbone produces a logit response map supervised in both probability and logit domains. LDMC constructs target--hard-negative pairs in the logit domain, ABS generates scale-adaptive annular penalty regions around targets, and False-Alarm Focal Loss reweights high-confidence background errors. All components are jointly optimized without additional inference overhead.}
    \label{fig:framework}
\end{figure*}

\subsection{Overall Architecture and Optimization Objective}

The overall training pipeline of our method is illustrated in Fig.~\ref{fig:framework}. This paper proposes a framework for discriminative feature learning and shape refinement specifically designed for infrared small target detection. Unlike prior works that focus on designing complex feature extraction modules, we aim to reconstruct the supervision signals during the training phase. By adding LDMC, ABS, and FAFL to the basic SLS loss of MSHNet, we enhance the model's capability to recognize small targets and suppress high-confidence false alarms.

We select MSHNet as the baseline detector because it adopts a relatively concise U-Net-based architecture with multi-scale prediction heads, making it a suitable and representative benchmark for infrared small target detection under significant target scale variations. More importantly, the structural simplicity of MSHNet allows the effectiveness of our method to be demonstrated more clearly, without being entangled with additional architectural complexity. Notably, our approach is a \textbf{plug-and-play} training strategy that introduces no extra inference parameters. During testing, the network architecture remains identical to the original MSHNet.

\textbf{Mathematical Formulation.}
Given an input infrared image $\mathcal{I} \in \mathbb{R}^{H \times W \times C}$ and its corresponding ground truth label $\mathcal{Y} \in \{0, 1\}^{H \times W}$, we define the end-to-end mapping of the network in two stages:
\begin{itemize}
    \item \textbf{Feature Mapping Stage:} The final convolutional layer of the network outputs the raw response map $\mathcal{Z} \in \mathbb{R}^{H \times W}$. We define $\mathcal{Z}$ as the prediction in the logit domain, which contains uncompressed feature confidence information.
    \item \textbf{Probability Activation Stage:} The response is mapped to the $[0, 1]$ interval via an activation function to obtain the final probability prediction $\mathcal{P}_{prob}=\mathcal{P} = \sigma(\mathcal{Z})$.
\end{itemize}
Traditional supervision methods compute loss only on $\mathcal{P}$, which often leads to gradient vanishing in the logit domain $\mathcal{Z}$. The core idea of this paper is to construct a \textbf{dual-domain supervision system} that simultaneously imposes constraints on both $\mathcal{Z}$ and $\mathcal{P}$.

\textbf{Composite Optimization Objective.}
To overcome the limitations of a single loss function, we construct a multi-dimensional objective function, AC-SLSIoU. Following the logic introduced above, AC-SLSIoU adds three losses to the basic SLS loss of MSHNet: Logit-Domain Margin Constraint, Adaptive Boundary Suppression, and False-Alarm Focal Loss. Online hard-negative selection serves as a sampling tool for LDMC:

\begin{equation}
\begin{split}
\mathcal{L}_{total} = \mathcal{L}_{base} &+ \lambda_{1}\mathcal{L}_{LDMC} + \lambda_{2}\mathcal{L}_{FAFL} \\
&+ \lambda_{3}\mathcal{L}_{boundary}
\end{split}
\end{equation}

The definitions for each term are as follows:
\begin{itemize}
    \item $\mathcal{L}_{base}(\mathcal{P}, \mathcal{Y})$: Adopts the SLS loss from MSHNet, a scale- and location-sensitive supervision term that reweights the IoU component according to target scale discrepancy and introduces a center-based location penalty, thereby helping the detector distinguish targets of different sizes and localize them more precisely.
    \item $\mathcal{L}_{LDMC}(\mathcal{Z}, \mathcal{Y})$: Logit-Domain Margin Constraint, which widens the response margin between targets and hard-negative background pixels in the logit domain.
    \item $\mathcal{L}_{FAFL}(\mathcal{P}, \mathcal{Y})$: False-Alarm Focal Loss, which assigns larger weights to high-probability negative samples and suppresses persistent false alarms.
    \item $\mathcal{L}_{boundary}(\mathcal{P}, \mathcal{Y})$: Adaptive Boundary Suppression loss, which generates scale-adaptive annular penalty regions around targets to eliminate prediction halos and refine contours.
\end{itemize}

In implementation, the selected hard negatives can also be assigned a lightweight BCE-style mining regularizer. This auxiliary term is used only to stabilize the suppression of selected clutter pixels and is not treated as an independent principal component of the proposed objective.

Through this multi-dimensional joint supervision, we guide the network to learn feature representations that possess both strong discriminative power and high localization precision.

\subsection{Logit-Domain Margin Constraint}

Infrared small target detection faces extreme foreground-background imbalance and gradient saturation issues within the probability domain. To address these challenges, we use LDMC as the main discriminative constraint and employ online hard-negative selection only to provide informative background samples for constructing target-clutter pairs.

To enhance the model's discriminative power for weak targets, we design a Logit-Domain Margin Constraint (LDMC) mechanism. The core idea of this mechanism stems from the ``pulling positive samples closer and pushing negative samples apart'' paradigm in contrastive learning.

\textbf{Evolution Logic from Contrastive Learning to Pairwise Ranking:} In standard contrastive learning, instance-level comparisons are typically performed in the feature space via the InfoNCE loss. However, since infrared small target datasets are characterized by an extremely small proportion of target pixels, such detection tasks face an extreme pixel-level imbalance in discrimination. Inspired by recent works such as SAIST and WeCoL, we argue that explicitly widening the feature distribution between targets and background through contrastive relationships is key to improving discriminability. Given that the probability domain suffers from severe gradient saturation when processing dim and small targets, we choose to place the contrastive operation in the unbounded logit domain. To concretize the abstract contrastive idea into an optimizable objective, we adopt a pairwise ranking logic: instead of independently constraining each pixel, we construct ``target-clutter'' pixel pairs and enforce a requirement that the positive pixel response $z^+$ must be higher than the negative pixel response $z^-$ by an explicit safe margin $m$.

The effectiveness of LDMC highly depends on the quality of the sample pairs. Since most background regions in infrared images produce extremely low logit values, random sampling of background pixels leads to low training efficiency and fails to provide effective gradients. Before constructing the logit-domain pairs, we identify hard negative pixels from the background region using a standard Online Hard Negative Mining (OHNM) strategy. Given the predicted probability map $\mathcal{P} = \sigma(\mathcal{Z})$ and the ground truth mask $\mathcal{Y}$, the background index set and corresponding probability responses are defined as $\mathcal{I}_{bg} = \{(i,j) \mid \mathcal{Y}_{i,j} = 0\}$ and $\mathcal{S}_{bg} = \{ \mathcal{P}_{i,j} \mid (i,j) \in \mathcal{I}_{bg} \}$, respectively. A quantile threshold is then computed as:
\begin{equation}
\tau_{hard} = \text{Quantile}(\mathcal{S}_{bg}, q)
\end{equation}
where $q$ is the quantile parameter. The hard negative index set is obtained by retaining background pixels with relatively high predicted confidence:
\begin{equation}
\mathcal{I}_{hard} = \{(i,j) \in \mathcal{I}_{bg} \mid \mathcal{P}_{i,j} \ge \tau_{hard}\}.
\end{equation}
Mapping these indices back to the logit response map gives the hard negative logit set $\Omega^-_{hard} = \{ \mathcal{Z}_{i,j} \mid (i,j) \in \mathcal{I}_{hard} \}$. These pixels correspond to clutter regions that are more likely to be confused with true targets. Therefore, we utilize the hard negative samples selected by this OHNM strategy to construct the most informative contrastive pairs. By enlarging the distance between positive samples and hard negatives, LDMC directly improves target-clutter separability in the logit domain.

For positive samples, we draw logits from the foreground target set $\mathcal{S}_{fg} = \{ \mathcal{Z}_{i,j} \mid \mathcal{Y}_{i,j}=1 \}$. To reduce pair imbalance, we randomly construct a balanced positive set $\Omega^+_{bal}$, ensuring its size matches that of the selected hard negatives: $|\Omega^+_{bal}| = \min(|\Omega^-_{hard}|, |\mathcal{S}_{fg}|)$.

As a minor auxiliary regularizer, the selected hard negatives can also be penalized by a BCE-style mining term:
\begin{equation}
\mathcal{L}_{mining} = - \frac{1}{|\mathcal{I}_{hard}|} \sum_{(i,j) \in \mathcal{I}_{hard}} \log(1 - \mathcal{P}_{i,j} + \epsilon).
\end{equation}
This term is only used to provide an additional suppression signal for the selected background pixels, while the principal discriminative effect is provided by the LDMC ranking constraint below.

\textbf{Loss Formulation and Derivation.}
Based on the balanced positive set $\Omega^+_{bal}$ and the hard negative set $\Omega^-_{hard}$, we construct the pair set $\mathcal{S}_{pair}$ in an all-to-all manner, i.e., each positive sample is paired with every hard negative sample:
\begin{equation}
\mathcal{S}_{pair} = \{(z^+, z^-) \mid z^+ \in \Omega^+_{bal}, z^- \in \Omega^-_{hard} \}
\end{equation}
Based on this pair set, our core objective is to satisfy the partial ordering relationship $z^+ - z^- \ge m$. To transform this hard constraint into a smooth and differentiable optimization objective, we introduce the Softplus function for relaxation. The Logit-Domain Margin Constraint loss $\mathcal{L}_{LDMC}$ is defined as follows:
\begin{equation}
\mathcal{L}_{LDMC} = \frac{1}{|\mathcal{S}_{pair}|} \sum_{(z^+, z^-) \in \mathcal{S}_{pair}} \ln \left( 1 + \exp \left( \frac{-(z^+ - z^- - m)}{\tau} \right) \right)
\end{equation}
where:
\begin{itemize}
    \item \textbf{Softplus Activation}: Serving as a smooth form of ranking loss, it provides a small but continuous push even when the constraint is partially satisfied (i.e., $z^+ - z^- \approx m$), preventing the model from premature convergence.
    \item \textbf{Margin Parameter $m$}: Defines the minimum ``safe distance'' between the target and clutter in the logit space, acting as a signal amplifier.
    \item \textbf{Temperature Coefficient $\tau$}: Used to adjust the sensitivity of the loss function to the response differences.
\end{itemize}

\textbf{Gradient Analysis.}
LDMC is essentially a ``response amplifier.'' Unlike probability-domain losses where gradients vanish rapidly as predicted values approach boundaries, the gradient of LDMC depends solely on the relative difference $\Delta z$ between positive and negative responses. As long as the score gap between the target and the background does not reach the predefined margin $m$, the mechanism generates a continuous gradient push, forcing the model to mine more discriminative features during the feature extraction stage. This feature enhancement mechanism, derived from ranking logic, is the core reason why LDMC can significantly improve $P_d$ while suppressing false alarms.

\begin{figure*}[!t]
    \centering
        \includegraphics[width=0.9\textwidth]{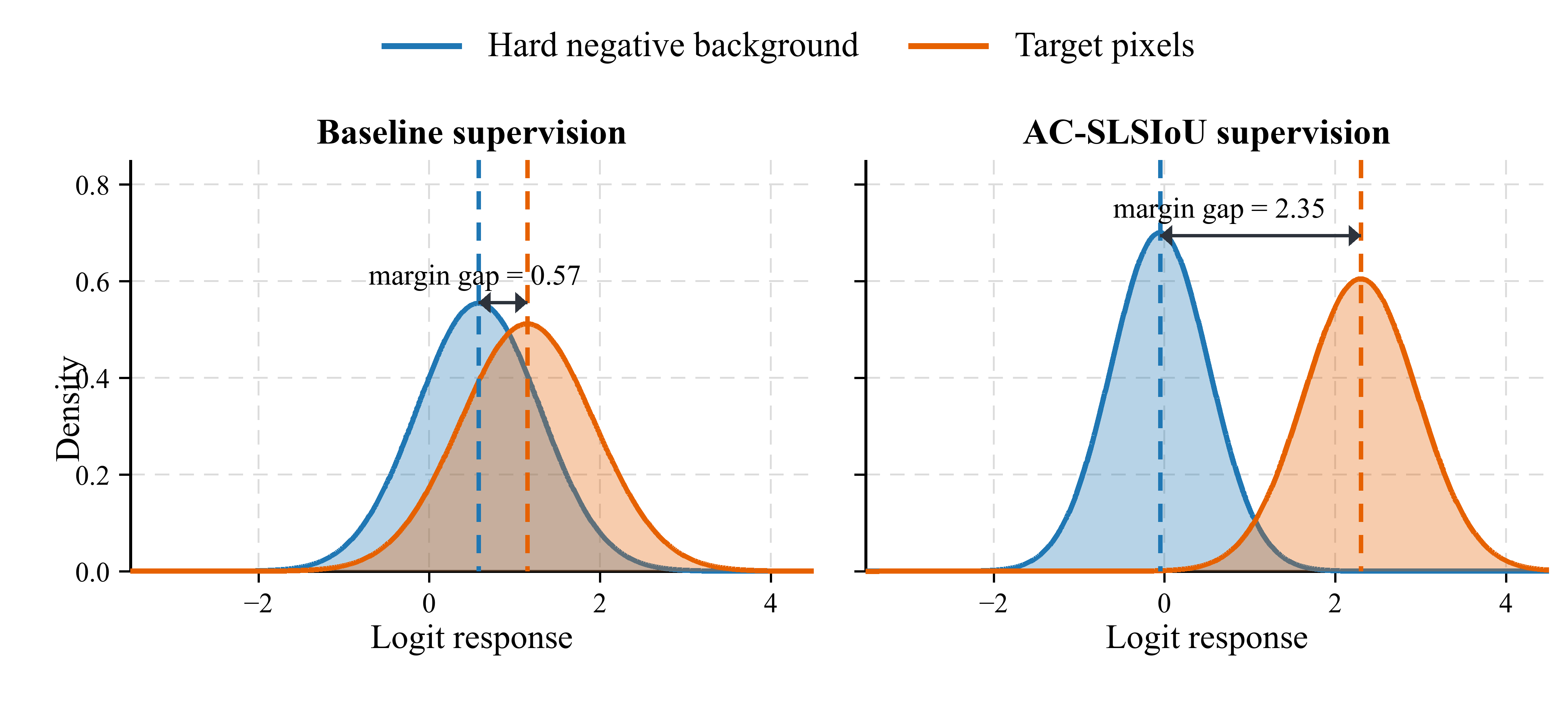}
    \caption{Logit distribution comparison between target pixels and hard negative background pixels under baseline supervision and AC-SLSIoU supervision. The proposed logit-domain constraint enlarges the response gap between target pixels and ambiguous background clutter.}
        \label{fig:logit_distribution}
\end{figure*}

To further validate the discriminative effect of LDMC, Fig.~\ref{fig:logit_distribution} compares the logit distributions of target pixels and hard negative background pixels. Under baseline probability-domain supervision, the two distributions are highly overlapping, indicating that weak targets and bright clutter remain difficult to separate in ambiguous infrared scenes. In contrast, AC-SLSIoU shifts target pixels toward higher logit responses while suppressing hard negative background pixels to lower responses. As a result, the average margin gap is enlarged from 0.57 to 2.35, which directly verifies that LDMC enhances target-background separability in the unbounded logit space and provides stronger optimization signals for weak target discrimination.

\subsection{Adaptive Boundary Suppression}

In IRSTD, accurately determining the precise boundaries of targets is often challenging due to the diffusion effect of thermal radiation, which leads to predictions exhibiting a blurred, halo-like appearance. This structural uncertainty not only degrades the Intersection over Union (IoU) but also hampers the compactness of target localization. To address this issue, we propose an Adaptive Boundary Suppression (ABS) mechanism, designed to dynamically constrain the predicted contours based on the scale of the targets.

Infrared small targets exhibit significant scale variations, typically ranging from $2 \times 2$ to $50 \times 50$ pixels. Utilizing a fixed radius for boundary suppression often results in ``over-suppression'' of tiny targets or ``under-constraint'' for larger ones.

To this end, we design an adaptive boundary suppression loss based on the target scale. First, we decompose the ground-truth mask $\mathcal{Y}$ into connected target components $\{\mathcal{Y}_c\}_{c=1}^{C}$. For each component, we dynamically calculate a component-specific dilation kernel $k_c$ based on its target area $Area(\mathcal{Y}_c)$:
\begin{equation}
k_c = \max\left(3, k_{base} + 2 \cdot \left\lfloor \min\left(\frac{\sqrt{Area(\mathcal{Y}_c)}}{s_k}, 2\right) \right\rfloor\right)
\end{equation}
where $k_{base}$ is the base kernel size (set to 3 by default) and $s_k$ is a scaling coefficient. This formula ensures that for larger target components, the dilation kernel automatically increases to generate a wider suppression band, while for extremely small target components, the kernel maintains a minimum size to avoid excessive suppression. The resulting component-specific kernels are used in the subsequent morphological dilation operation to expand each target region in the ground truth mask, thereby constructing scale-adaptive boundary bands around the targets as the regions to be penalized.

Based on the component-specific kernel $k_c$, we generate the boundary penalty mask $\mathcal{M}_{bound}$. We perform morphological dilation on each connected target component $\mathcal{Y}_c$ using an all-ones kernel of size $k_c \times k_c$, and then merge all component-level annular regions:
\begin{equation}
\mathcal{M}_{bound} =
\bigcup_{c=1}^{C}
\left(
\operatorname{Dilate}(\mathcal{Y}_c, k_c) \odot (1-\mathcal{Y})
\right)
\end{equation}
where $\odot$ denotes element-wise multiplication. This produces annular regions that precisely cover the background immediately surrounding each target while excluding all ground-truth target pixels. 

The final boundary loss is defined as the mean value of the predicted probabilities $\mathcal{P}_{prob}$ within these regions, forcing the network to suppress overflowed predictions back within the target boundaries:
\begin{equation}
\mathcal{L}_{boundary} = \frac{\sum_{(i,j)} \mathcal{P}_{prob}^{(i,j)} \cdot \mathcal{M}_{bound}^{(i,j)}}{\sum_{(i,j)} \mathcal{M}_{bound}^{(i,j)} + \epsilon}
\end{equation}

Through this adaptive pruning strategy, the model can significantly eliminate the halo effect and achieve pixel-level shape refinement. Meanwhile, tighter boundary constraints implicitly promote the accuracy of target center localization.

\subsection{False-Alarm Focal Loss}
Even after employing LDMC and ABS, numerous stubborn false alarms may still remain in complex infrared scenes. These responses usually correspond to background interference regions that obtain exceedingly high prediction probabilities over the positive class. Inspired by Focal Loss, which emphasizes hard samples through probability-dependent weighting, we introduce False-Alarm Focal Loss to assign larger penalties to negative samples with high predicted probabilities.

This term does not directly reuse the hard-negative set selected by OHNM. Instead, it is activated by a probability-domain confidence threshold over all background pixels. Thus, OHNM provides quantile-based hard samples for LDMC, whereas False-Alarm Focal Loss penalizes all background pixels whose predicted probabilities exceed $\tau_{conf}$.

We first define a high-confidence mask $\mathcal{M}_{high}$, which activates only those background pixels whose predicted probabilities exceed a specific threshold $\tau_{conf}$:
\begin{equation}
\mathcal{M}_{high} = \mathbb{I}(\mathcal{P} > \tau_{conf}) \cdot (1 - \mathcal{Y})
\end{equation}
For these high-risk regions, we apply a dynamic weighting penalty in the form of Focal Loss. Since these are negative samples (Label=0), the Focal weight term is $\mathcal{P}^\gamma$:
\begin{equation}
\mathcal{L}_{focal} = - \frac{1}{N_{high}} \sum_{(i,j)} \left[ \alpha \cdot (\mathcal{P}_{i,j})^\gamma \cdot \log(1 - \mathcal{P}_{i,j}) \cdot \mathcal{M}_{high}^{(i,j)} \right]
\end{equation}
where $\alpha$ is the balancing factor and $\gamma$ is the focusing parameter. The parameter $\gamma$ controls the strength of focusing on high-confidence false alarms: as $\gamma$ increases, the weights of low-confidence background pixels are further suppressed, while background pixels with high predicted probabilities still retain large penalty weights. Therefore, this term concentrates optimization on the background interference regions that are more difficult to suppress. Unlike the standard Focal Loss, our implementation introduces a hard threshold via $\mathcal{M}_{high}$, meaning that the loss is only activated when the confidence of the false alarm reaches the predefined threshold. This design ensures that the model imposes stronger penalty gradients on the most persistent false alarms, thereby effectively suppressing high-score false detections.

\section{Experiments}

\subsection{Experimental Setup}

\textbf{Datasets.} 
To ensure a fair comparison with the baseline method MSHNet, we conducted experiments on two widely used public datasets: IRSTD-1k and NUDT-SIRST.
\begin{itemize}
    \item \textbf{IRSTD-1k} contains 1,001 infrared images with rich background clutter. Following the standard protocol, we divided the dataset into training and testing sets with a ratio of 4:1.
    \item \textbf{NUDT-SIRST} consists of 1,327 images covering a diverse range of target scales. This dataset is equally split into training and testing sets with a 1:1 ratio.
\end{itemize}

\textbf{Evaluation Metrics.} 
We employ three standard metrics to comprehensively evaluate the detection performance: Intersection over Union (\textbf{IoU}), Probability of Detection ($P_d$), and False Alarm Rate ($F_a$).
\begin{itemize}
    \item \textbf{IoU} measures the pixel-level overlap between the predicted mask and the ground truth mask, reflecting the precision of shape segmentation.
    \item \textbf{$P_d$} measures the ratio of correctly detected targets to the total number of actual targets, reflecting the model's capability to capture weak targets.
    \item \textbf{$F_a$} measures the ratio of falsely predicted target pixels to the total number of pixels in the image, reflecting the model's ability to suppress background noise. Following common IRSTD evaluation practice, $F_a$ is reported in units of $10^{-6}$.
\end{itemize}

\textbf{Implementation Details.} 
The proposed method is implemented based on the PyTorch framework. All input images are resized to a resolution of $256 \times 256$. The network is trained using the AdaGrad optimizer with an initial learning rate of 0.05 and a batch size of 4. The model is trained for a total of 400 epochs. All experiments were conducted on a single NVIDIA GeForce RTX 4090 GPU.

The loss hyperparameters are fixed as follows. For the composite objective, the weights of $\mathcal{L}_{LDMC}$, $\mathcal{L}_{focal}$, and $\mathcal{L}_{boundary}$ are set to $\lambda_1=0.038$, $\lambda_2=0.38$, and $\lambda_3=0.019$, respectively. For LDMC, the OHNM quantile, logit-domain margin, and Softplus ranking temperature are set to $q=0.95$, $m=0.12$, and $\tau=1.0$, respectively. For False-Alarm Focal Loss, the high-confidence threshold, balancing factor, and focusing parameter are set to $\tau_{conf}=0.60$, $\alpha=0.25$, and $\gamma=2.5$, respectively. For ABS, the base dilation kernel and scale coefficient are set to $k_{base}=3$ and $s_k=8.0$, respectively.
\subsection{Comparison with State-of-the-Arts}

We compare AC-SLSIoU with representative traditional and deep-learning-based IRSTD methods on the IRSTD-1k and NUDT-SIRST datasets. The traditional baselines include filtering-based methods (Top-Hat and Max-Median), local-contrast-based methods (WSLCM and TLLCM), and low-rank-based methods (PSTNN and MSLSTIPT). The deep learning baselines include classic CNN-based detectors (ISNet, DNANet), the scale-and-location-sensitive baseline (MSHNet), and three recent advanced approaches from 2024--2025: MIRSAM, a multimodal framework that adapts the Segment Anything Model with vision-language cues for infrared scenes; TDA, which introduces a target-driven adaptive loss to enhance detection on small and low-contrast targets; and PConv + SDLoss, which combines pinwheel-shaped convolution with scale-based dynamic loss for improved feature extraction and training stability.

\subsubsection{Overall Quantitative Performance}
As shown in Table~\ref{tab:sota_comparison_full}, AC-SLSIoU achieves the best performance across all six metrics on both datasets. The corresponding performance distribution is visualized in Fig.~\ref{fig:performance_distribution}, highlighting the overall advantage of AC-SLSIoU among the compared methods. On IRSTD-1k, it attains an $IoU$ of 71.55\%, a $P_d$ of 95.88\%, and an $F_a$ of 5.54, outperforming the second-best method MIRSAM by $+1.98\%$ in $IoU$ and $+1.24\%$ in $P_d$, while reducing $F_a$ by $3.13$. On NUDT-SIRST, AC-SLSIoU reaches an $IoU$ of 84.47\%, a $P_d$ of 98.56\%, and an $F_a$ of 4.91, consistently surpassing all traditional and deep-learning-based competitors with a clear margin. These results demonstrate that the proposed logit-domain margin constraint, adaptive boundary suppression, and False-Alarm Focal Loss provide a unified and effective solution for both IRSTD-1k and NUDT-SIRST benchmark scenarios.

\subsubsection{Comparison with Traditional and Classic CNN-Based Methods}
Traditional methods rely on handcrafted priors and therefore exhibit limited robustness in complex infrared scenes. Filtering-based and local-contrast-based methods suffer from either low overlap accuracy or extremely high false alarm rates; for example, WSLCM and TLLCM obtain $P_d$ values above 70\% on IRSTD-1k, but their $F_a$ values reach 6619 and 6738, respectively. Low-rank-based methods improve background modeling to some extent, yet their IoU values remain far below those of deep models. These results indicate that manually designed priors are insufficient for handling severe clutter, low contrast, and target scale variations.

ISNet and DNANet are two representative early deep architectures widely cited in the IRSTD literature. ISNet employs shape-aware constraints to guide detection; however, as shown in Table~\ref{tab:sota_comparison_full}, its $F_a$ reaches 27.92 on IRSTD-1k and 34.65 on NUDT-SIRST, indicating that shape priors alone are insufficient for suppressing complex background clutter. DNANet improves feature representation through dense nested interactions and achieves a competitive $IoU$ of 79.98\% on NUDT-SIRST. Nevertheless, its IRSTD-1k $IoU$ (65.71\%) remains substantially lower than AC-SLSIoU (71.55\%), and its $F_a$ on IRSTD-1k (17.61) is more than three times that of our method (5.54). This gap reveals that architectural advancements, while beneficial for multi-scale feature aggregation, cannot fully overcome the gradient saturation problem inherent in probability-domain supervision—a limitation that our logit-domain reformulation directly addresses.

\subsubsection{Comparison with Recent Advanced Methods}
MIRSAM, TDA, and PConv + SDLoss represent the latest design philosophies in IRSTD. MIRSAM achieves the closest overall performance to AC-SLSIoU, particularly on IRSTD-1k ($IoU$ 69.57\% vs.\ 71.55\%, $F_a$ 8.67 vs.\ 5.54), benefiting from the powerful visual priors of the SAM foundation model and supplementary language guidance. However, its reliance on multimodal inputs (infrared image paired with textual descriptions) restricts applicability in deployment scenarios where such annotations are impractical. TDA enhances the SLS loss through patch-based target-driven adaptive weighting and achieves an IRSTD-1k $IoU$ of 68.45\% and $P_d$ of 93.17\%. However, its NUDT-SIRST $F_a$ remains at 20.65, substantially higher than that of AC-SLSIoU (4.91), suggesting that per-target adaptive reweighting may overfit to the specific target characteristics of real images and compromise false alarm suppression on data with diverse target scales. PConv + SDLoss improves bottom-layer feature extraction via Gaussian-aligned pinwheel convolutions and stabilizes training with scale-based dynamic loss, yielding a well-balanced $F_a$ (9.67 on IRSTD-1k, 12.52 on NUDT-SIRST). However, its $IoU$ (67.12\% on IRSTD-1k, 75.17\% on NUDT-SIRST) trails AC-SLSIoU by over 4\% and 9\%, respectively, indicating that convolution-level and loss-weighting enhancements, while beneficial, cannot substitute for the explicit discriminative constraints that operate in the unbounded logit space.

\subsubsection{Improvement over the Baseline Model (MSHNet)}
As a plug-and-play training strategy applied to the MSHNet architecture, AC-SLSIoU delivers consistent and substantial improvements over the baseline trained with the basic SLS loss. On IRSTD-1k, $IoU$ increases to 71.55\% from 67.16\% ($+4.39\%$), $P_d$ improves to 95.88\% from 93.88\% ($+2.00\%$), and $F_a$ is reduced to 5.54 from 15.03, corresponding to a 63.1\% reduction in false alarms. On NUDT-SIRST, $IoU$ rises to 84.47\% from 80.55\% ($+3.92\%$), $P_d$ increases to 98.56\% from 97.99\% ($+0.57\%$), and $F_a$ drops to 4.91 from 11.77, a 58.3\% reduction. These improvements confirm that the three added losses--LDMC, ABS, and False-Alarm Focal Loss--effectively enhance the original MSHNet supervision without increasing inference cost.

\begin{table*}[!t]
    \centering
    \caption{Quantitative comparison with representative traditional and deep-learning-based methods on IRSTD-1k and NUDT-SIRST datasets. The best results are highlighted in \textbf{bold}. Higher IoU and $P_d$ are better, while lower $F_a$ is better.}
    \label{tab:sota_comparison_full}
    \resizebox{\textwidth}{!}{
    \begin{tabular}{c|c|ccc|ccc}
        \toprule
        \multirow{2}{*}{\textbf{Method}} & \multirow{2}{*}{\textbf{Description}} & \multicolumn{3}{c|}{\textbf{IRSTD-1k}} & \multicolumn{3}{c}{\textbf{NUDT-SIRST}} \\
        \cmidrule(lr){3-5} \cmidrule(lr){6-8}
         & & IoU (\%) & $P_d$ (\%) & $F_a$ ($10^{-6}$) & IoU (\%) & $P_d$ (\%) & $F_a$ ($10^{-6}$) \\
         \midrule
        Top-Hat & \multirow{2}{*}{Filtering} & 10.06 & 75.11 & 1432 & 20.72 & 78.41 & 166.7 \\
        Max-Median & & 6.998 & 65.21 & 59.73 & 4.197 & 58.41 & 36.89 \\
        \midrule
        WSLCM & \multirow{2}{*}{Local Contrast} & 3.452 & 72.44 & 6619 & 2.283 & 56.82 & 1309 \\
        TLLCM & & 3.311 & 77.39 & 6738 & 2.176 & 62.01 & 1608 \\
         \midrule
        PSTNN & \multirow{2}{*}{Low Rank} & 24.57 & 71.99 & 35.26 & 14.85 & 66.13 & 44.17 \\
        MSLSTIPT & & 11.43 & 79.03 & 1524 & 8.342 & 47.40 & 888.1 \\
         \midrule
        ISNet & \multirow{7}{*}{Deep Learning} & 62.88 & 92.59 & 27.92 & 67.86 & 92.59 & 34.65 \\
        MIRSAM & & 69.57 & 94.64 & 8.67 & 80.16 & 97.56 & 10.12 \\
        TDA  & & 68.45 & 93.17 & 12.45 & 77.42 & 97.63 & 20.65 \\
        PConv + SDLoss & & 67.12 & 92.34 & 9.67 & 75.17 & 96.54 & 12.52 \\
        DNANet & & 65.71 & 91.84 & 17.61 & 79.98 & 96.93 & 12.78 \\
        MSHNet (Baseline) & & 67.16 & 93.88 & 15.03 & 80.55 & 97.99 & 11.77 \\
        \textbf{AC-SLSIoU (Ours)} & & \textbf{71.55} & \textbf{95.88} & \textbf{5.54} & \textbf{84.47}& \textbf{98.56}& \textbf{4.91}\\
        \bottomrule
    \end{tabular}
    }
\end{table*}

\begin{figure*}[!t]
  \centering
    \includegraphics[width=\textwidth]{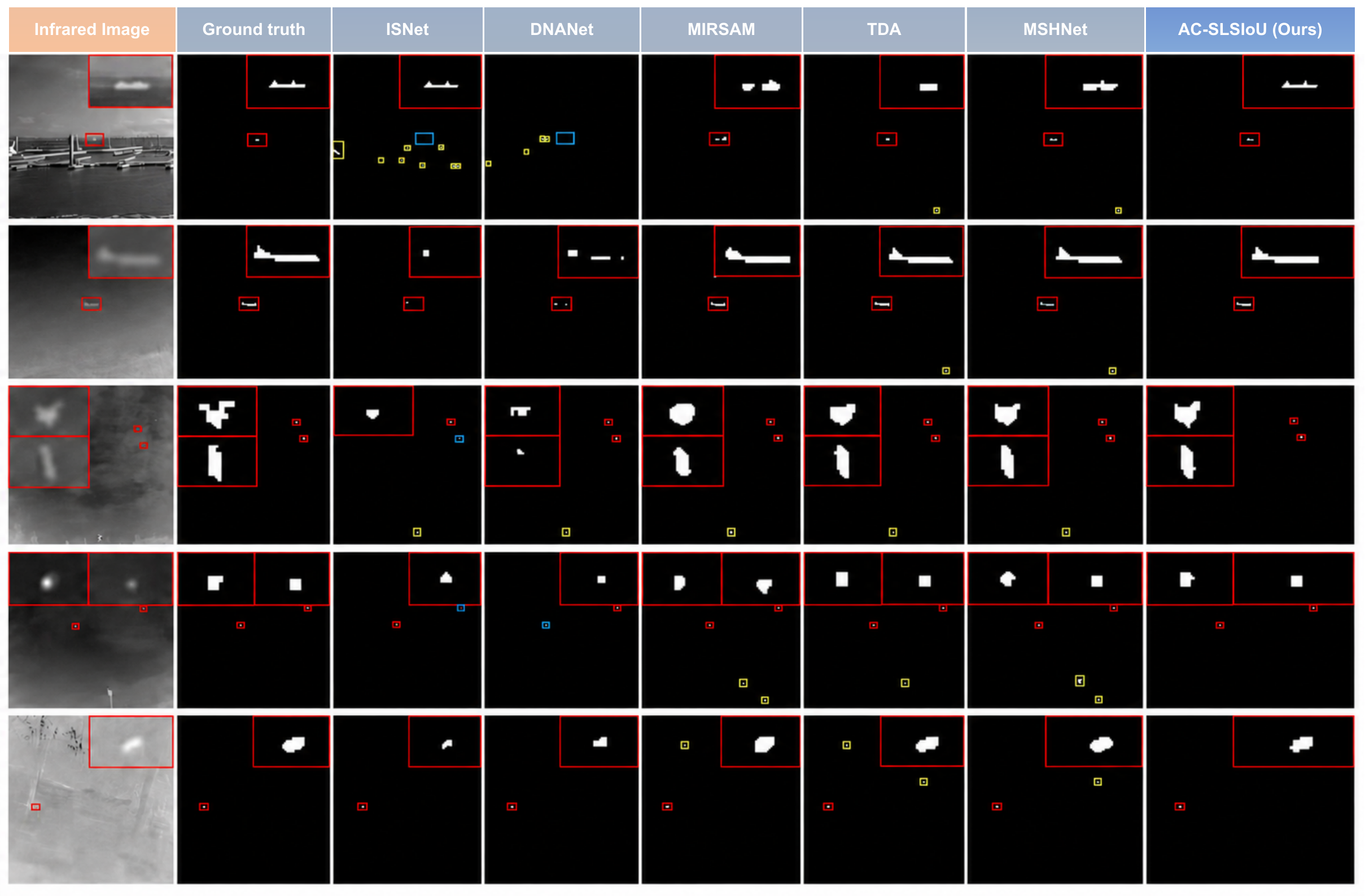} 
    \caption{Visual comparison of different IRSTD methods. Red boxes indicate target regions and enlarged local details, blue boxes indicate missed detections, and yellow boxes indicate false alarms.}
    \label{fig:visual_comparison}
\end{figure*}

Fig.~\ref{fig:visual_comparison} provides a qualitative comparison of different methods under representative infrared scenes. In this figure, red boxes indicate target regions and enlarged local details, blue boxes denote missed detections, and yellow boxes denote false alarms. Compared with ISNet, DNANet, MIRSAM, TDA, and the MSHNet baseline, AC-SLSIoU produces more compact target masks and fewer scattered false responses in cluttered background regions. The enlarged target regions show that the proposed loss preserves weak target responses while reducing halo-like boundary overflow. These visual results are consistent with the quantitative improvements in Table~\ref{tab:sota_comparison_full}, especially the lower $F_a$ and higher IoU achieved by AC-SLSIoU.

\subsection{Ablation Study}

To verify the effectiveness of the core components in the AC-SLSIoU loss function, we conducted ablation experiments on IRSTD-1k and NUDT-SIRST. Using MSHNet trained with the basic SLS loss as the baseline, we evaluate each component from two complementary perspectives. First, each proposed module is individually added to the baseline to examine its independent effect. Second, each module is removed from the complete AC-SLSIoU objective to analyze its role within the joint optimization framework. The quantitative results are summarized in Table~\ref{tab:ablation_strict}.

Compared with the MSHNet baseline, the complete AC-SLSIoU loss achieves consistent overall improvements on both datasets. On IRSTD-1k, IoU increases from 67.16\% to 71.55\%, $P_d$ improves from 93.88\% to 95.88\%, and $F_a$ decreases from 15.03 to 5.54, corresponding to a 63.1\% reduction in false alarms. On NUDT-SIRST, IoU rises from 80.55\% to 84.47\%, $P_d$ increases from 97.99\% to 98.56\%, and $F_a$ is reduced from 11.77 to 4.91, corresponding to a 58.3\% reduction. These results demonstrate that the joint use of LDMC, ABS, and False-Alarm Focal Loss improves segmentation accuracy, target detection, and clutter suppression simultaneously.

The single-component variants show that no individual term fully accounts for the final improvement. LDMC alone improves IoU on both datasets and substantially lowers false alarms, reducing $F_a$ from 15.03 to 7.14 on IRSTD-1k and from 11.77 to 5.86 on NUDT-SIRST, but its $P_d$ is slightly lower than the baseline. This indicates that logit-domain separation is effective for clutter suppression but still needs complementary constraints to preserve target sensitivity. ABS alone improves $P_d$ to 95.20\% on IRSTD-1k and 98.41\% on NUDT-SIRST, but its $F_a$ increases, suggesting that boundary refinement by itself tends to retain more target-like responses unless paired with discriminative suppression. False-Alarm Focal Loss alone provides the strongest standalone IoU gains, reaching 69.74\% and 82.36\% on the two datasets, and also reduces $F_a$ to 8.73 and 7.12. However, it does not achieve the best $P_d$, which shows that reweighting high-probability negative samples is useful but insufficient as the only constraint.

The removal experiments further clarify the complementary roles of the three main components in the complete framework. Removing ABS lowers IoU from 71.55\% to 66.93\% on IRSTD-1k and from 84.47\% to 79.88\% on NUDT-SIRST, while $F_a$ rises sharply to 19.74 and 17.94, respectively. This severe degradation confirms that adaptive boundary suppression is essential for keeping the discriminative responses compact and preventing boundary overflow. Removing False-Alarm Focal Loss reduces IoU to 69.41\% and 83.06\%, and increases $F_a$ to 13.36 and 10.76, even though this variant obtains the highest $P_d$ on both datasets. Therefore, this focal loss mainly converts higher detection sensitivity into a more reliable detection result by suppressing persistent high-confidence false alarms. Finally, removing LDMC lowers IoU to 68.38\% and 82.12\%, reduces $P_d$ to 93.50\% and 96.71\%, and causes $F_a$ to rise to 19.13 and 16.28. This confirms that the logit-domain margin constraint remains the core discriminative driver of AC-SLSIoU.

Overall, the ablation results show that the proposed modules are complementary rather than simply additive. LDMC establishes the target-clutter margin, ABS constrains the spatial extent of the response, and False-Alarm Focal Loss penalizes persistent false alarms by assigning larger weights to high-probability negative samples. The complete AC-SLSIoU objective achieves the best IoU and the lowest $F_a$ on both datasets, while maintaining a $P_d$ close to the best removal variant. This balance indicates that the full objective provides the most reliable trade-off among segmentation accuracy, target detection, and clutter suppression.

\begin{table*}[!t]
    \centering
    \caption{Ablation study of AC-SLSIoU components on IRSTD-1k and NUDT-SIRST. Checkmarks indicate enabled components. LDMC, ABS, and FAFL denote Logit-Domain Margin Constraint, Adaptive Boundary Suppression, and False-Alarm Focal Loss, respectively. The best results are highlighted in \textbf{bold}.}
    \label{tab:ablation_strict}
    \begingroup
    \setlength{\tabcolsep}{4.5pt}
    \renewcommand{\arraystretch}{1.08}
    \resizebox{\textwidth}{!}{%
    \begin{tabular}{@{}cccccccccc@{}}
        \toprule
        \textbf{Base} & \textbf{LDMC} & \textbf{ABS} & \textbf{FAFL} &
        \multicolumn{3}{c}{\textbf{IRSTD-1k}} &
        \multicolumn{3}{c}{\textbf{NUDT-SIRST}} \\
        \cmidrule(lr){5-7} \cmidrule(l){8-10}
        & & & & IoU (\%) & $P_d$ (\%) & $F_a$ ($10^{-6}$) & IoU (\%) & $P_d$ (\%) & $F_a$ ($10^{-6}$) \\
        \midrule
        \cmark &        &        &        & 67.16 & 93.88 & 15.03 & 80.55 & 97.99 & 11.77 \\
        \cmark & \cmark &        &        & 68.34 & 93.16 & 7.14 & 81.73 & 96.92 & 5.86 \\
        \cmark &        & \cmark &        & 67.86 & 95.20 & 15.56 & 81.18 & 98.41 & 13.27 \\
        \cmark &        &        & \cmark & 69.74 & 94.52 & 8.73 & 82.36 & 97.74 & 7.12 \\
        \cmark & \cmark &        & \cmark & 66.93 & 94.18 & 19.74 & 79.88 & 97.63 & 17.94 \\
        \cmark & \cmark & \cmark &        & 69.41 & \textbf{96.22} & 13.36 & 83.06 & \textbf{98.82} & 10.76 \\
        \cmark &        & \cmark & \cmark & 68.38 & 93.50 & 19.13 & 82.12 & 96.71 & 16.28 \\
        \rowcolor{yellow!18}
        \cmark & \cmark & \cmark & \cmark & \textbf{71.55} & 95.88 & \textbf{5.54} & \textbf{84.47} & 98.56 & \textbf{4.91} \\
        \bottomrule
    \end{tabular}%
    }
    \endgroup
\end{table*}

\subsection{Generalization Analysis across Different Backbones}

To further evaluate the universality and "plug-and-play" capability of the proposed AC-SLSIoU loss function, we conducted extensive experiments on several representative architectures beyond the main MSHNet setting. These include HDNet~\cite{xu2025hdnet}, which introduces hybrid spatial-frequency enhancement, DNANet, which features multi-scale feature fusion, and the recent PConv-based detector, which employs pinwheel-shaped convolution to enhance infrared small target feature extraction.

\begin{table}[!t]
\centering
\caption{Performance comparison before and after applying the AC-SLSIoU loss function on the IRSTD-1k dataset. Higher IoU and $P_d$ are better, while lower $F_a$ is better.}
\label{tab:generalization}
\renewcommand{\arraystretch}{1.2}
\begin{tabular}{l|ccc}
\hline
\multirow{2}{*}{Method} & \multicolumn{3}{c}{IRSTD-1k} \\ \cline{2-4} 
 & IoU (\%) & $P_d$ (\%) & $F_a$ ($10^{-6}$) \\ \hline
MSHNet (Baseline) & 67.16 & 93.88 & 15.03 \\
\textbf{MSHNet + AC-SLSIoU } & \textbf{71.55} & \textbf{95.88} & \textbf{5.54} \\ \hline
HDNet & 70.26 & 94.56 & 4.33 \\
\textbf{HDNet + AC-SLSIoU } & \textbf{73.53} & \textbf{95.67} & \textbf{3.29} \\ \hline
DNANet & 65.71 & 91.84 & \textbf{17.61} \\
\textbf{DNANet + AC-SLSIoU } & \textbf{67.54} & \textbf{92.13} & 18.23 \\ \hline
PConv + SDLoss & 67.12 & 92.34 & 9.67 \\
\textbf{PConv + AC-SLSIoU } & \textbf{69.80} & \textbf{94.05} & \textbf{7.45} \\ \hline
\end{tabular}
\end{table}

As shown in Table \ref{tab:generalization}, the completed results on MSHNet, HDNet, DNANet, and PConv demonstrate that our proposed strategy consistently improves IoU across different backbones. The detailed experimental analysis is as follows:

\begin{itemize}
    \item \textbf{Adaptability to Hybrid-Domain Architectures}: On HDNet, which already combines spatial-domain feature extraction with frequency-domain high-frequency enhancement, AC-SLSIoU further improves IoU from 70.26\% to 73.53\% and $P_d$ from 94.56\% to 95.67\%, while reducing $F_a$ from 4.33 to 3.29. This corresponds to a +3.27\% IoU gain and a 24.0\% reduction in false alarms. These results indicate that the proposed logit-domain supervision remains effective even when the backbone has already incorporated strong frequency-based target enhancement.
    \item \textbf{Generalization to Multi-scale Architectures}: On the DNANet architecture, IoU improves from 65.71\% to 67.54\% (+1.83\%). Although $F_a$ showed a slight increase, the dual growth in $P_d$ and IoU suggests that AC-SLSIoU assists the model in locking onto extremely weak targets during the multi-scale feature aggregation process.
    \item \textbf{Extension to Recent PConv-based Detectors}: PConv + SDLoss is further introduced as a recent 2025 representative method. With AC-SLSIoU, PConv reaches 69.80\% IoU, 94.05\% $P_d$, and 7.45 $F_a$, improving IoU by +2.68\% and reducing false alarms by 23.0\% over PConv + SDLoss under the same training and evaluation protocol.
\end{itemize}

In summary, the completed cross-backbone results validate that by adding LDMC, ABS, and False-Alarm Focal Loss to the basic supervision objective, AC-SLSIoU effectively improves infrared small target detection across different backbones. Regardless of whether the underlying feature extraction paradigm emphasizes multi-scale aggregation, spatial-frequency enhancement, or specialized convolutional operators, our method consistently enhances the detection of infrared small targets. The HDNet result is reported as a cross-backbone validation rather than the main MSHNet-based comparison setting, further demonstrating the plug-and-play nature of the proposed loss.

\section{Conclusion}

In this paper, to address the issues of gradient vanishing caused by activation function saturation and the blurred edges of prediction results in IRSTD, we proposed an enhanced discriminative and shape-aware loss function, AC-SLSIoU.

First, by shifting the optimization focus from the restricted probability domain to the unbounded logit domain, this paper introduces a logit-domain pairwise contrastive constraint. This mechanism effectively amplifies the signal response of weak targets and overcomes the gradient saturation bottleneck in extremely low SNR scenarios. Second, targeting the halo effect caused by thermal diffusion, we designed an adaptive boundary suppression mechanism that dynamically adjusts the penalty ring according to the target scale, achieving pixel-level shape refinement and pruning. Finally, False-Alarm Focal Loss assigns larger weights to high-probability negative samples, further enhancing the model's ability to suppress persistent false alarms.

Experimental results on the IRSTD-1k and NUDT-SIRST datasets demonstrate that AC-SLSIoU significantly improves the detection rate $P_d$ and the Intersection over Union (IoU) of the model. Furthermore, generalization analysis across different backbones validates the universality and efficiency of this strategy as a plug-and-play training scheme. In summary, this research provides a new perspective for discriminative feature learning in infrared small target detection, holding substantial academic value and potential for engineering applications.

\end{document}